\documentclass{article}

\PassOptionsToPackage{numbers, compress}{natbib}

\usepackage[final]{neurips}
\usepackage[utf8]{inputenc} 
\usepackage[T1]{fontenc}    
\usepackage[colorlinks,linkcolor=red,anchorcolor=blue,citecolor=green]{hyperref}
\usepackage{url}            
\usepackage{booktabs}       
\usepackage{amsfonts}       
\usepackage{nicefrac}       
\usepackage{microtype}      
\usepackage{xcolor}         
\usepackage{multirow}
\usepackage{times}
\usepackage{epsfig}
\usepackage{subfloat}
\usepackage{eso-pic}
\usepackage{xspace}
\usepackage{tikz}
\usepackage{bbm}
\usepackage{colortbl,booktabs}
\usepackage{makecell}
\usepackage{diagbox} 
\usepackage{multicol}
\usepackage{array}
\usepackage{makecell}
\usepackage{amsmath, bm}
\usepackage{mathtools}
\usepackage{amsthm}
\usepackage{dsfont}
\usepackage{amssymb}
\usepackage{algorithm}
\usepackage{algpseudocode}
\usepackage{varwidth}
\usepackage{pifont}
\usepackage{makecell}
\usepackage{wrapfig}
\usepackage{enumitem}
\usepackage{caption}
\usepackage{listings}
\usepackage{color}
\usepackage{tikz}
\usetikzlibrary{arrows.meta}
\usepackage{graphicx}
\usepackage{tabularx}
\usepackage{siunitx}
\usepackage{subcaption, multirow, overpic, textpos}

\definecolor{green}{RGB}{0,150,10}
\definecolor{blue}{RGB}{0,148,181}

\title{EC-Diff: Fast and High-Quality Edge-Cloud Collaborative Inference for Diffusion Models}

\author{Jiajian Xie$^1$, Shengyu Zhang$^{1,*}$, Zhou Zhao$^1$, Fan Wu$^2$, Fei Wu$^1$\\
[1mm]
$^1$Zhejiang University $^2$Shanghai Jiao Tong University\\
\texttt{\small {xiejiajian@zju.edu.cn, sy\_zhang@zju.edu.cn}}
}

\begin{document}

\renewcommand{\thefootnote}{\fnsymbol{footnote}}
\footnotetext{*Corresponding author.}

\maketitle
\begin{abstract}
    Diffusion Models have shown remarkable proficiency in image and video synthesis. As model size and latency increase limit user experience, hybrid edge-cloud collaborative framework was recently proposed to realize fast inference and high-quality generation, where the cloud model initiates high-quality semantic planning and the edge model expedites later-stage refinement. However, excessive cloud denoising prolongs inference time, while insufficient steps cause semantic ambiguity, leading to inconsistency in edge model output. To address these challenges, we propose EC-Diff that accelerates cloud inference through gradient-based noise estimation while identifying the optimal point for cloud-edge handoff to maintain generation quality. Specifically, we design a K-step noise approximation strategy to reduce cloud inference frequency by using noise gradients between steps and applying cloud inference periodically to adjust errors. Then we design a two-stage greedy search algorithm to efficiently find the optimal parameters for noise approximation and edge model switching. Extensive experiments demonstrate that our method significantly enhances generation quality compared to edge inference, while achieving up to an average $2\times$ speedup in inference compared to cloud inference. Video samples and source code are available at \url{https://ec-diff.github.io/}. 
\end{abstract}

\section{Introduction}
\vspace{-0.15cm}

Diffusion models ~\citep{balaji2022ediff,ho2020denoising,nichol2021improved,rombach2022high,ho2022video} have demonstrated remarkable capabilities in high-quality image and video generation, finding extensive applications across various domains such as artistic creation and data augmentation. However, pursuing higher generation quality typically corresponds with larger model sizes, and the step-by-step denoising procedure further exacerbates computational intensity, resulting in significant latency. For example, SDXL-base ~\citep{podell2023sdxl} with 3.5B parameters requires approximately 10 seconds to generate a high-quality image of 1024$\times$1024 resolutions, while CogVideoX ~\citep{yang2024cogvideox} with 5B parameters demands approximately 4 minutes to produce a 49 frames video with 720×480 resolutions. Diffusion acceleration as an effective technique has been deeply explored recently, including progressive distillation ~\citep{salimans2022progressive,lin2024sdxl} and consistency models ~\citep{Song2023ConsistencyM,luo2023latent,wang2023videolcm} to reduce sampling steps, as well as feature reuse ~\citep{ma2024deepcache,ye2024training,zhao2024real} to minimize computational overhead. Despite these advances, large-scale model structures still require cloud-based inference implementations with high-end GPUs, imposing significant overhead on cloud servers as user requests grow exponentially. Although recent works on pruning and distillation of diffusion models ~\citep{fang2024structural,li2023snapfusion,kim2024bk} have enabled lightweight model deployment on the edge to reduce latency, these approaches often experience noticeable degradation in generation quality due to model simplification. Consequently, developing methods that simultaneously achieve high generation quality and low latency remains a critical challenge.

\begin{figure*}[t]
    \centering
    \resizebox{0.99\linewidth}{!}{
    \includegraphics{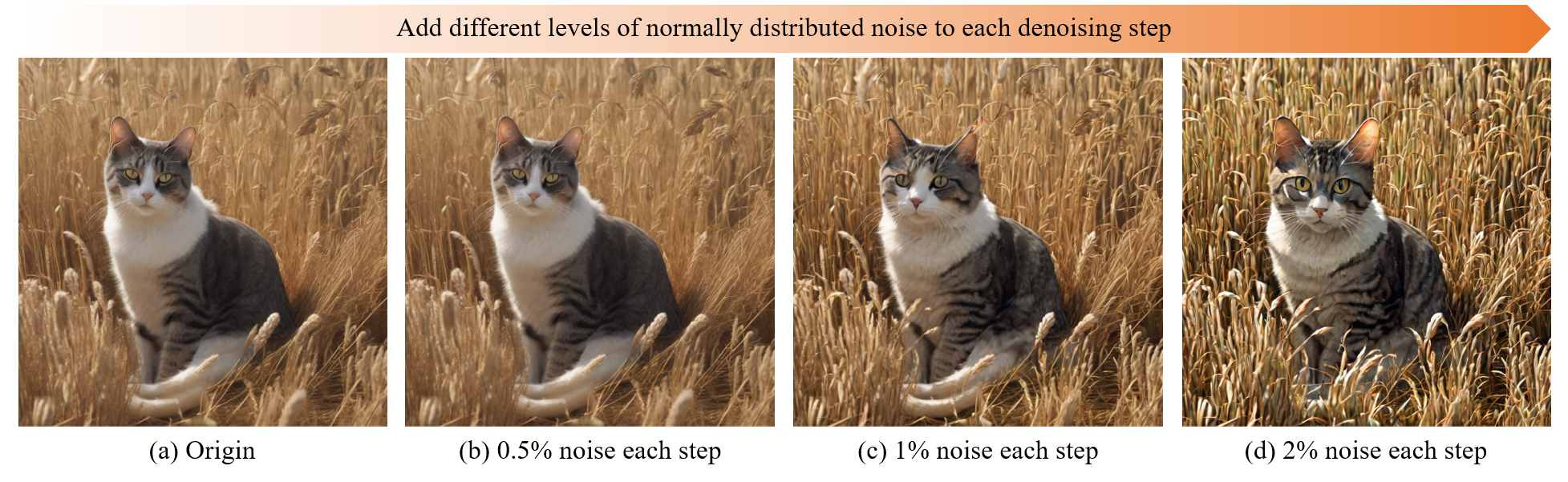}}
    \caption{\small Diffusion model exhibits a certain degree of denoising error adjustment capability. Adding 0.5\% normal distribution noise as denoising error to the latent at each step results in the same output as the original. However, with error ratios of 1\% and 2\%, quality degradation occurs, indicating the error exceeds the model's correction capacity.}
    \label{fig:error_correct}
\end{figure*}

More recently, HybridSD ~\citep{yan2024hybrid} introduced the first edge-cloud collaborative diffusion generated framework, which leverages the complementary strengths of high-quality cloud models and efficient edge models to achieve fast generation without compromising output quality. Specifically, the cloud model performs the predetermined number of denoising steps to establish clear semantic layouts in the latent space. The denoising process is then completed by the edge model, which primarily serves to refine the visual quality of the generated output. However, ensuring semantic clarity during the cloud denoising phase typically requires a substantial number of inference steps, significantly increasing overall latency. Conversely, insufficient cloud inference leads to ambiguous semantic layout information in the latent representation, causing the edge model's output to diverge from the intended semantic structure and resulting in quality degradation. This raises two critical challenges: (1) how to minimize cloud denoising time while preserving semantic structural clarity; (2) how to determine the optimal handoff point to edge-model processing to maximize inference speed without compromising generation quality.

To address these challenges, we propose a novel edge–cloud collaborative diffusion framework called EC-Diff. For cloud model acceleration, we observe that diffusion models possess inherent error-correction capabilities. As shown in ~\autoref{fig:error_correct}, when 0.5\% Gaussian noise is introduced to the latent representation after each denoising step, the final generated results remain virtually identical to the unperturbed outputs. However, as noise magnitude increases beyond the model's correcting threshold, generation quality degrades proportionally. Motivated by this observation, we introduce a $k$-step noise approximation strategy that accelerates cloud model inference while preserving semantic clarity by ensuring that error accumulation from approximated noise remains within the model's correction capacity. 
Specifically, we compute the noise gradient by analyzing the pattern of noise changes between consecutive time steps. The gradient is then utilzed to approximated the model-predicted noise for the subsequent $k$ steps in the denoising process. Since the error between the approximated noise and the model-predicted noise accumulates and amplifies during the denoising process, we utilize the model's error correction capablity to rectify accumulated errors and update the noise gradient, initiating a new cycle of strategy loops. 
For the decision of the optimal cloud-edge switching point, since the error accumulation caused by the $k$-step approximation strategy on the cloud impacts the performance of the edge model, we designed a two-stage greedy search algorithm optimizing a quality-latency objective function. The first stage identifies optimal parameter combinations for the k-step approximation strategy on the cloud side, and the second stage fixes these parameters and locates the most effective cloud-edge switching point, balancing computational efficiency with output quality.

The main contributions of this paper are summarized as follows:
\begin{itemize}[leftmargin=*]
	\item We propose a $k$-step noise approximation strategy that computes approximation noise across multiple steps using noise gradients to accelerate cloud model inference, while leveraging the model's error correction capability to mitigate cumulative errors introduced by the approximation process.
	\item We design a two-stage greedy search algorithm to determine both the optimal parameter combinations for the $k$-step approximation strategy and the optimal cloud-edge switching point.
	\item Extensive experiments conducted on both image and video diffusion models demonstrate that our method can achieve up to a 2.5$\times$ speedup with better preservation quality.
\end{itemize}
\section{Related Works}
\subsection{Diffusion Models}

Diffusion models \cite{ho2020denoising,rombach2022high} have achieved great success in generative tasks, surpassing GANs \cite{goodfellow2020generative,li2020gan} and autoregressive models \cite{parmar2018image,ding2021cogview} through their superior generation quality and diversity. Early diffusion models \cite{ho2020denoising,song2020denoising} implemented iterative denoising processes directly in pixel space, which imposing substantial computational burdens. To improve efficiency, latent diffusion models (LDMs) \cite{rombach2022high} project images into lower-dimensional latent space before executing forward and reverse diffusion processes, which further evolves into Stable Diffusion (SD) family \cite{podell2023sdxl,sauer2024fast}. Recently, video diffusion models \cite{blattmann2023align,blattmann2023stable,wang2023modelscope,yang2024cogvideox,deng2023mv,sun2024mm} have attracted increasing attention. VideoLDM \cite{blattmann2023align} extend 2D-UNet into 3D-UNet by injecting temporal layers and operate a latent denoising process to save computational resources. CogVideoX \cite{yang2024cogvideox} introduces an expert transformer with adaptive LayerNorm to improve the spatial and temporal coherence of the generated video. Despite the high quality achieved by diffusion models, the inherent nature of the reverse process which needs high computational cost slows down the inference process.

\subsection{Accelerating Diffusion Models Inference}

Current works accelerate diffusion models inference can be divided into the following aspects. (1) Reducing the number of sampling steps \cite{salimans2022progressive,lin2024sdxl,sauer2024adversarial,Song2023ConsistencyM,luo2023latent,wang2023videolcm,lyu2022accelerating,moon2023early,huang2022prodiff}. Progressive distillation \cite{salimans2022progressive,lin2024sdxl} involves gradually distilling a high-step diffusion model to generate images with fewer steps. Consistency models \cite{Song2023ConsistencyM,luo2023latent} train a model to maintain consistency across different time points, enabling one-step image generation for faster results. Besides, some works \cite{lyu2022accelerating,moon2023early} introduce early stop mechanism into diffusion models. (2) Feature reuse \cite{ma2024deepcache,wimbauer2024cache,ye2024training,zhao2024real,deng2023efficiency,tian2024qvd}. DeepCache \cite{ma2024deepcache} notices the feature redundancy in the denoising process and introduces a cache mechanism to reuse pre-computed features. AdaptiveDiffusion \cite{ye2024training} develops a skipping strategy to decide whether the noise prediction should be inferred or reused from the previous noise.

Despite these methods are efficient for inference, the large model structure limits edge application. Although lightweight models \cite{fang2024structural,li2023snapfusion,kim2024bk} are edge-deployment friendly, significant quality degradation persists due to their inherently limited computational capacity. More recently, HybridSD \cite{yan2024hybrid} proposes an edge-cloud collaboration framework to address these problems, where the cloud model initiates high-quality semantic planning and the device model expedites later-stage refinement. However, excessive cloud denoising steps prolong inference time, while insufficient cloud denoising leaves semantic layouts in the latent space ambiguously defined, causing device model to deviate from the cloud’s generative intent and degrade output consistency. 

In this paper, we propose a gradient-based noise approximation strategy to accelerate cloud inference, while exploring the optimal point for cloud-edge handoff to maintain the generation quality with minimum time cost.

\section{The Proposed Approach}

\begin{figure*}[ht]
    \centering
    \includegraphics[width=1\textwidth]{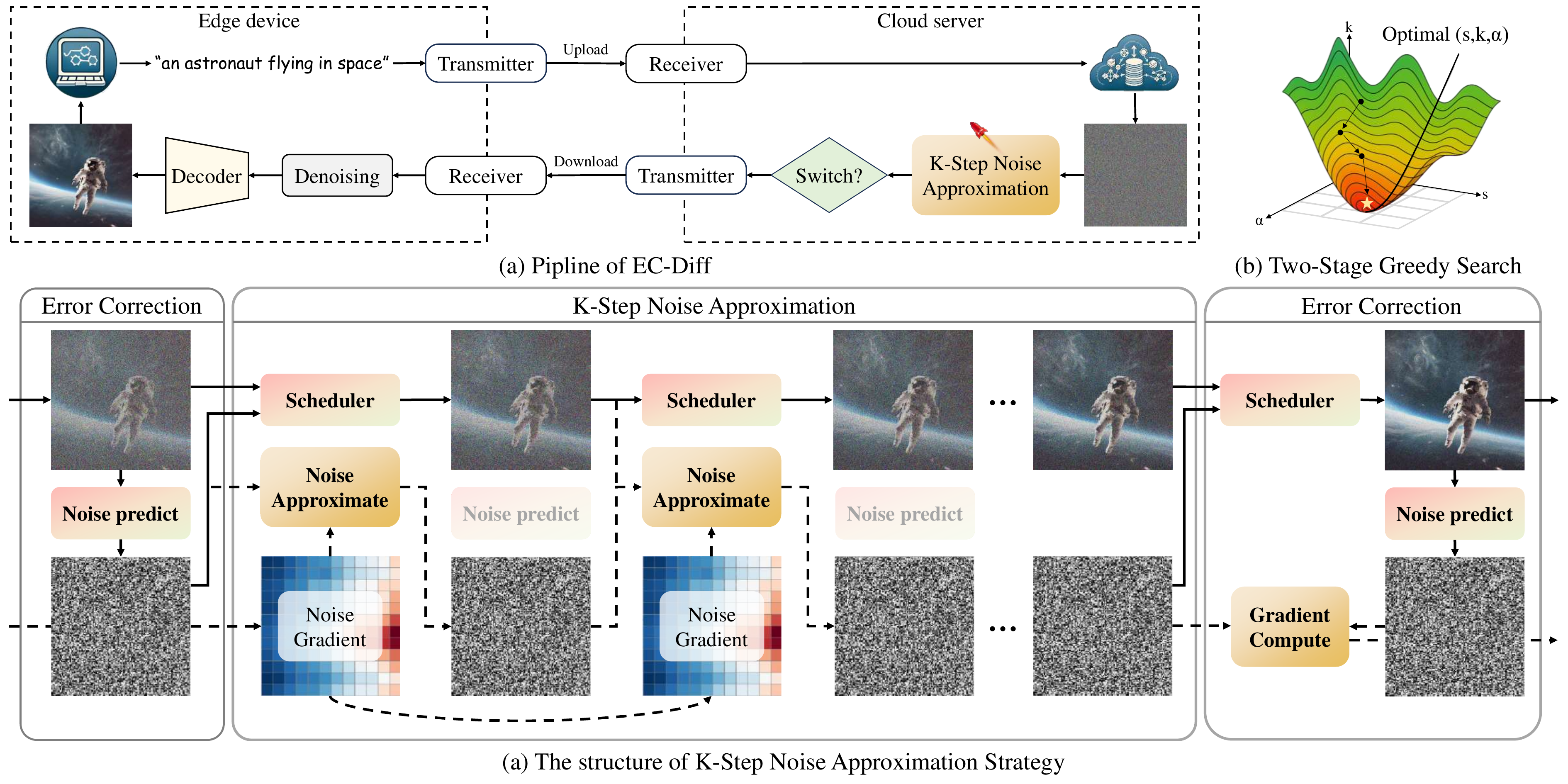}
    \caption{Overview of the proposed EC-Diff. Subfigure (a) depicts the collaborative architecture where the cloud model accelerates denoising for several steps before switching to the edge model for the remaining inference. Subfigure (b) depicts a two-stage greedy search to find the optimal parameter combination, where $s$ is the switching point, $k$ is the number of noise approximation steps, and $\alpha$ is the smoothing factor. Subfigure (c) depicts the cyclic process of accelerated inference using $k$-step noise approximation with error correction.}
    \label{fig:method}
\end{figure*}

In this section, we provide a detailed description of the architectural components of our proposed framework, EC-Diff. As shown in \autoref{fig:method}, the overall framework consists of two main modules: 1) K-Step Noise Approximation Strategy, which accelerates cloud inference and guarantees semantic clarity; 2) Two-Stage Greedy Search, which balances output quality with inference efficiency to determine both the optimal parameter combinations for the $k$-step approximation strategy and the optimal cloud-edge switching point. The following subsections detail the design of each module.

\vspace{-0.20cm}
\subsection{Preliminary}
\vspace{-0.15cm}

\noindent{\textbf{Diffusion Denoising Process.}} Diffusion models consist of a forward process that progressively corrupts data with Gaussian noise and a reverse process that reconstructs the clean data through iterative denoising. During inference, only the reverse process is executed, starting from the Gaussian noise $x_T\sim \mathcal{N}(0,I)$ and progressively refines $x_T$ into the target output $x_0$ conditioned on input signal $c$ (e.g., text prompts), where $T$ is the predefined number of denoising steps. The refinement follows a generic update rule defined at each timestep $t$ :
\begin{equation}\label{eq:eq_1}
	x_{t-1}=f(t-1) \cdot x_{t}-g(t-1) \cdot \epsilon_{\theta}(x_{t}, t, c),\,\,\,\,\,\,\,t=1,\ldots,T, 
\end{equation}
where $\epsilon_{\theta}(x_{t}, t, c)$ is a noise prediction network (e.g., UNet \cite{ronneberger2015u}) trained to estimate the noise component in $x_t$ by taking $x_t$, timestep $t$ and an additional condition $c$ as input, while $f(t)$ and $g(t)$ are coefficients determined by the sampler (a.k.a, scheduler) \cite{ho2020denoising,song2020denoising}. Both the noise prediction network $\epsilon_{\theta}$ and sampler coefficients $f(t)$, $g(t)$ directly impact generation quality. In the edge-cloud collaborative framework, using the same sampler ensures the consistency of $f(t)$ and $g(t)$, so the generation quality is primarily affected by the differences of the noise prediction model $\epsilon_{\theta}$.

\noindent\textbf{Noise Approximation Error.} Previous work \cite{ye2024training} have proven that noise predictions between consecutive denoising steps exhibit reusable patterns. However, as visualized in \autoref{fig:pre_inference}, the model-predicted noise exhibits variations across different time steps. When noise is reused without proper adjustment, errors occur between the reused noise and the model-predicted noise. These errors accumulate and amplify throughout the denoising process, particularly affecting edge-cloud collaborative frameworks by compromising the clarity of semantic planning in the cloud-generated latent, which subsequently impacts edge model generation quality. To effectively minimize accumulated errors, we design a k-step gradient-based noise approximation strategy during the cloud inference phase, which will be proven the cumulative error remains sufficiently small to be corrected by the model's inherent corrective capabilities in Sec.5. Given a certain timestep $i$ to use approximated noise, the error accumulation can be formulated using \autoref{eq:eq_2}:
\begin{equation}\label{eq:eq_2}
	\begin{aligned}
		x_{i-1}^{'} &= f(i-1) \cdot x_{i}-g(i-1) \cdot (\epsilon_\theta(x_i, i,c)-\beta_i)
		\\
		&= x_{i-1}+g(i-1)\cdot \beta_i,
	\end{aligned}
\end{equation}
where $\beta_{i}$ is the error between the approximated noise and the model-predicted noise $\epsilon_\theta(x_i, i,c)$ in timestep $i$.

\subsection{K-Step Noise Approximation Strategy}

\subsubsection{\textbf{Noise Gradient Initialization.}}
\vspace{-0.15cm}

\begin{wrapfigure}{r}{0.5\textwidth}
  \centering
  \resizebox{\linewidth}{!}{
	\begin{subfigure}{0.5\textwidth}
		\includegraphics[width=\textwidth]{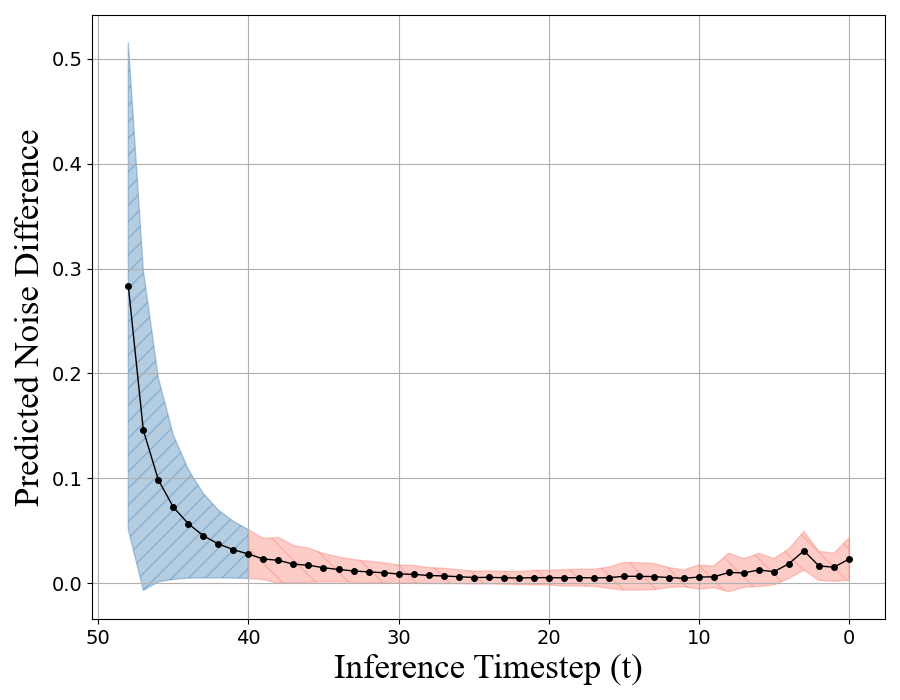}
		\subcaption{}
		\label{fig:pre_inference_pic}
	\end{subfigure}
	\begin{subfigure}{0.5\textwidth}
		\includegraphics[width=\textwidth]{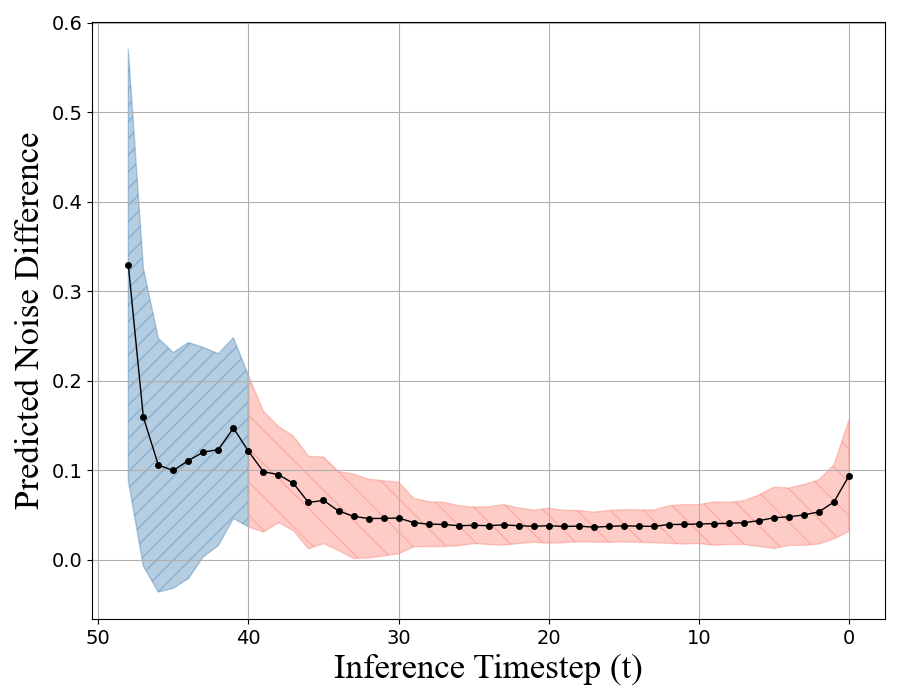}
		\subcaption{}
		\label{fig:pre_inference_video}
	\end{subfigure}}
  \caption{Difference in predicted noise between two consecutive inference steps.  Each data point in the figure is an average of all noise differences, while the shaded area indicates the variance. (a) text-to-image generation task. (b) text-to-video generation task.}
  \label{fig:pre_inference}
\end{wrapfigure}

Previous work \cite{liu2024faster} characterizes the denoising steps by semantics-planning and fidelity-improving stages, indicating that cross-attention maps exhibit substantial fluctuations during the semantic planning phase and remain relatively stable throughout the fidelity refinement phase. In alignment with these findings, we observe a similar pattern when analyzing the difference in predicted noise between two consecutive steps. \autoref{fig:pre_inference_pic} shows the noise difference using SDXL-base \cite{podell2023sdxl} performs text-to-image generation on the MS-COCO dataset \cite{lin2014microsoft}, while \autoref{fig:pre_inference_video} shows the noise difference for text-to-video generation using CogVideoX \cite{yang2024cogvideox} on VBench \cite{huang2024vbench}. Both image and video generation tasks demonstrate pronounced fluctuations in noise differences during the semantic planning phase, followed by a substantially more stable pattern in subsequent phases. Motivated by this observation, we allow the cloud model to perform several denoising steps until the predicted noise differences exhibit a stable trend before initializing the noise gradient computation. Let $i$ denote the number of cloud model pre-inference steps, the initial noise gradient can be formulated as:
\begin{equation}\label{eq:eq_3}
	\delta \epsilon = \epsilon_\theta(x_i, i, c) - \epsilon_\theta(x_{i+1}, i+1, c) .
\end{equation}

\subsubsection{\textbf{K-Step Noise Approximation.}}
\label{sec:k-step}
Following the initialization of the noise gradient, the denoising process over the next $k$ steps is executed using approximated noise, without requiring inference from the cloud model. Specifically, we assume that the noise over these k steps follows a stable and linear trend. Consequently, we can approximate the model-predicted noise at each step by applying the noise gradient to the previous step's noise. Based on \autoref{eq:eq_3}, the approximate noise for step $i-2$ can be formulated as:
\begin{equation}\label{eq:eq_4}
	\epsilon_\theta^{'}(x_{i-1}, {i-1}, c) = \epsilon_\theta(x_i, i, c) + \delta \epsilon.
\end{equation}

Similarly, we extend this process to the $k$-th step and obtain the following formulation:
\begin{equation}\label{eq:eq_5}
	\epsilon_\theta^{'}(x_{i-1-k}, {i-1-k}, c) = \epsilon_\theta(x_i, i, c) + k \times \delta \epsilon.
\end{equation}

Since the discrepancy between the approximated noise and the model-predicted noise leads to error accumulation, we use the cloud model to perform a single inference step after completing $k$ steps of sampling based on \autoref{eq:eq_1}, thereby leveraging the model’s corrective capability to mitigate this accumulate error. 

We then update the noise gradient using the difference between the model-predicted noise and the approximated noise at the k-th step. However, as shown in ~\autoref{fig:proof_2} in Appendix, this difference often exceeds the true gradient since the model tends to predict more noise to compensate for the accumulated approximation error. As a result, the updated noise gradient may be biased with a larger magnitude. To address this, we introduce a smoothing factor $\alpha$ in the gradient update step using the following formula:
\begin{equation}\label{eq:eq_6}
	\delta \epsilon = \alpha \times (\epsilon_\theta(x_{i-k-2}, i-k-2, c) - \epsilon_\theta^{'}(x_{i-1-k}, i-1-k, c) ).
\end{equation}

The process then iterates with a new round of k-step noise approximation followed by model correction, continuing this alternating sequence until the edge model takes over the inference process. The complete $k$-step noise approximation strategy is summarized in Algorithm \autoref{alg:k-step} in Appendix.

\subsubsection{\textbf{Error Estimation.}}
To validate the effectiveness of our k-step approximation strategy, we theoretically analyze the upper bound of the error between the original output and approximated output. We use the form of \autoref{eq:eq_2} to represent the error accumulation.
\\ 
\textbf{Theorem 1.} Assume that the cloud model performs $i$ inference steps to obtain $x_{i-1}$ , then uses $k$ steps of approximated noise to obtain $x_{i-k-1}$, and applies cloud model inference for error correction to obtain $x_{i-k-2}$, then the following in-equation holds:
\begin{equation}\label{eq:eq_7}
	\begin{aligned}
		\Delta x_{i-k-2} = O(\sum\limits_{m = 0}^{{\rm{k - }}1} {{\beta _{i - 1 - m}}}  - \beta _{i - k - 1}^{'}).
	\end{aligned}
\end{equation}\textbf{}

The proof can be found in \ref{sec:first_apply_kstep} in Appendix. From \autoref{eq:eq_7}, it can be observed that the upper bound of the error in the first application of the strategy depends on whether the model's error correction capability can compensate for the accumulated deviation between the noise gradient and the true direction of noise change.
\\
\textbf{Theorem 2.} Assume that the latent obtained after the first application of the strategy is $x_{i-1}$, the error accumulation after every $k$-step noise approximation in subsequent strategy cycles follows the in-equation below:
\begin{equation}\label{eq:eq_8}
	\Delta x_{i-1-k} =  O(\sum\limits_{m = 1}^k {({\beta^{\triangle}_{i - m}} - (1 + m\alpha )\beta _i^{'} - m\alpha {\beta_{i + 1}})}).
\end{equation}
The proof can be found in \ref{sec:second_apply_kstep} in Appendix. Assuming the difference between the origin model-predicted noise $\epsilon_\theta(x_{i}, {i}, c)$ and $\epsilon_\theta(x_{i+1}, {i+1}, c)$ is used as the noise gradient $\delta \epsilon_{\text{org}}$, the resulting error $\beta^{\triangle}_{i - m}$ is consistent with the meaning of $\beta$ in \autoref{eq:eq_7}. However, due to the influence of the previous strategy cycle, the updated noise gradient contains two additional terms,  $\beta_{i+1}$ and $\beta^{'}_{i}$, causing a discrepancy between the new noise gradient and $\delta \epsilon_{\text{org}}$. Therefore, the final error accumulation depends on whether the new noise gradient can approach $\beta^{\triangle}_{i - m}$ under the influence of the smoothing factor $\alpha$. Meanwhile, when the accumulated error $\Delta x_{i-1-k}$ exceeds the model's error correction capacity, the remaining error will persist into the next cycle.

Therefore, when the gradient smoothing factor $\alpha$, the number of approximations $k$, and the number of cloud model pre-inference steps are optimally combined, the accumulated error will remain within the model's error correction capacity, enabling nearly quality-lossless fast generation. In the next section, we will introduce the optimization method to find this combination.

\subsection{Two-Stage Greedy Search Algorithm}
In addition to the parameters within the $k$-step approximation strategy that influence edge model performance, the timing of the switching to edge model inference represents a critical factor. Therefore, our goal is to determine the optimal combination of pre-inference steps $p$, approximated noise steps $k$, gradient smoothing factor $\alpha$, and edge model switching point $s$.

\subsubsection{\textbf{Formulation of the optimization objective.}} Our goal is to generate results approximating those generated by the cloud model, while minimizing inference time to approach that of the edge model. For generation quality assessment, we employ SSIM\cite{wang2004image} to compare the generated results between the cloud model and our method. Let $x^{\text{cloud}}$ denote the generation of the cloud model and $x^{\text{ec-diff}}$ represent the generation of our method, the quality evaluation metric is formulated as:
\begin{equation}\label{eq:eq_9}
	\text{Quality} = SSIM(x_i^{\text{cloud}},x_i^{\text{ec-diff}}).
\end{equation}

For inference time efficiency, we calculate the ratio of the improvement in inference time using our method compared to the cloud model's inference time. Let $t^{\text{cloud}}$ denote the inference time of the cloud model, $t^{\text{edge}}$ denote the inference time of the edge model, $t^{\text{ec-diff}}$ denote the inference time of our method, the inference time efficiency evaluation metric is formulated as:
\begin{equation}\label{eq:eq_10}
	\text{Efficiency} = \frac{{t_i^{\text{cloud}} - t_i^{\text{ec-diff}}}}{{t_i^{\text{cloud}} - t_i^{\text{edge}}}} .
\end{equation}

To minimize computational burden on cloud servers, we formulate a cloud load objective function. Let $s^{'}$ represent the pre-defined earliest cloud-edge switching point and $T$ denotes the total inference steps, the metric is formulated as :
\begin{equation}\label{eq:eq_10}
	\text{Burden} = \frac{{T - s}}{{T - s^{'}}} .
\end{equation}

Since both metrics range from 0 to 1, we obtain our optimization objective by the weighted sum of these two goals:
\begin{equation}\label{eq:eq_11}
	\text{Objective} = w_{1}\times \text{Quality}+w_{2}\times \text{Efficiency}+w_{3}\times \text{Burden}.
\end{equation}

\subsubsection{\textbf{Two-Stage Greedy Search.}} The parameter $p$ is selected based on the noise fluctuation results presented in \autoref{fig:pre_inference}. Then the first stage seeks the optimal $(k,\alpha)$ combination for the $k$-step noise approximation strategy used in the cloud model, ensuring the best generation quality and speed with minimal error accumulation, thus minimizing its impact on the cloud-edge switching point. The second stage solves for the optimal cloud-edge switching point given the best $(k,\alpha)$ combination. Both stages use a greedy algorithm with an early stopping mechanism. Specifically, we track the impact of parameter changes on the objective function. If the change is beneficial to the objective function, we continue exploring in that direction; otherwise, we explore in the opposite direction. If no improvement is observed after three consecutive parameter adjustments, we consider the optimal solution found and terminate the search. The complete two-stage greedy search algorithm is summarized in Algorithm \autoref{alg:greedy_1} and Algorithm \autoref{alg:greedy_2} in Appendix.

\begin{table*}[t]
	\caption{Quantitative results of T2I and T2V tasks. The best and second-best results are in bold and underlined.}  
	\label{tab:quantitive}
    \resizebox{1\linewidth}{!}
    {
	\begin{tabularx}{1.325\linewidth}{>{\centering\arraybackslash}c >{\centering\arraybackslash}c >{\centering\arraybackslash}c >{\centering\arraybackslash}c >{\centering\arraybackslash}c >{\centering\arraybackslash}c >{\centering\arraybackslash}c >{\centering\arraybackslash}c}
		\textbf{Model} & \textbf{Methods} & \textbf{PSNR↑} & \textbf{LPIPS↓} & \textbf{SSIM↑} & \textbf{VBench↑} & \textbf{Latency (s)↓} & \textbf{Speedup↑} \\
		\toprule
		\multirow{3}{*}{SD-v1.4 + BK-SDM-Small} & HybridSD (k=10) & \textbf{29.508} & \textbf{0.035} & \textbf{0.923} & --- & 1.61 & 1.05$\times$\\
		& HybridSD (k=25) & 23.121 & 0.134 & 0.786 & --- & \underline{1.39} & \underline{1.23$\times$} \\
		& \cellcolor{gray!15}\textbf{Ours} & \cellcolor{gray!15}\underline{29.402} & \cellcolor{gray!15}\underline{0.040} & \cellcolor{gray!15}\underline{0.919} & \cellcolor{gray!15}--- & \cellcolor{gray!15}\textbf{1.16} & \cellcolor{gray!15}\textbf{1.47$\times$}\\
		\midrule
		\multirow{3}{*}{SD-v1.4 + BK-SDM-Tiny} & HybridSD (k=10) & \textbf{29.389} & \textbf{0.036} & \textbf{0.919} & --- & 1.58 & 1.08$\times$\\
		& HybridSD (k=25) & 22.811 & 0.141 & 0.781 & --- & \underline{1.36} & \underline{1.25$\times$}\\
		& \cellcolor{gray!15}\textbf{Ours} & \cellcolor{gray!15}\underline{29.363} & \cellcolor{gray!15}\underline{0.049} & \cellcolor{gray!15}\underline{0.912} & \cellcolor{gray!15}--- & \cellcolor{gray!15}\textbf{1.05} & \cellcolor{gray!15}\textbf{1.62$\times$}\\
		\midrule
		\multirow{3}{*}{SDXL-base + SSD-1B} & HybridSD (k=10) & \underline{30.588} & \textbf{0.046} & \textbf{0.909} & --- & 5.87 & 1.07$\times$\\
		& HybridSD (k=25) & 25.201 & 0.132 & 0.821 & --- & \underline{5.28} & \underline{1.19$\times$}\\
		& \cellcolor{gray!15}\textbf{Ours} & \cellcolor{gray!15}\textbf{30.610} & \cellcolor{gray!15}\underline{0.061} & \cellcolor{gray!15}\underline{0.905} & \cellcolor{gray!15}--- & \cellcolor{gray!15}\textbf{2.77} & \cellcolor{gray!15}\textbf{2.26$\times$}\\
		\midrule
		\multirow{3}{*}{SDXL-base + SSD-vega} & HybridSD (k=10) & \textbf{29.322} & \textbf{0.058} & \textbf{0.889} & --- & 5.64 & 1.11$\times$\\
		& HybridSD (k=25) & 24.029 & 0.157 & 0.792 & --- & \underline{4.76} & \underline{1.32$\times$}\\
		& \cellcolor{gray!15}\textbf{Ours} & \cellcolor{gray!15}\underline{28.542} & \cellcolor{gray!15}\underline{0.063} & \cellcolor{gray!15}\underline{0.870} & \cellcolor{gray!15}--- & \cellcolor{gray!15}\textbf{2.61} & \cellcolor{gray!15}\textbf{2.40$\times$}\\
		\midrule
		\multirow{3}{*}{CogvideoX-5B + CogvideoX-2B} & HybridSD (k=10) & \textbf{28.733} & \textbf{0.055} & \underline{0.895} & \textbf{81.57} & 211.17 & 1.16$\times$\\
		& HybridSD (k=25) & 19.684 & 0.178 & 0.780 & 80.57 & 165.95 & 1.48$\times$\\
		& \cellcolor{gray!15}\textbf{Ours} & \cellcolor{gray!15}\underline{28.478} & \cellcolor{gray!15}\underline{0.069} & \cellcolor{gray!15}\textbf{0.898} & \cellcolor{gray!15}\underline{81.54} & \cellcolor{gray!15} \textbf{105.61} & \cellcolor{gray!15}\textbf{2.31$\times$}\\
	\end{tabularx}
    }
\end{table*}
\vspace{-0.2cm}
\section{Experiments}

\vspace{-0.1cm}
\subsection{Experimental Setup}

\noindent{\textbf{Base Models and Comparison Baseline.}} We conduct experiments on text-to-image (T2I) and text-to-video (T2V) generation tasks. For the T2I generation task, we utilize SD-v1.4 \cite{hf-sdv1-4} and SDXL-base \cite{podell2023sdxl} as cloud models. Specifically, for SD-v1.4, we select the BK-SDM-Small and BK-SDM-Tiny \cite{kim2024bk} as edge models, while for SDXL-base, we choose the SSD-1B and SSD-vega \cite{gupta2024progressive} as edge models. For the T2V generation task, we employ the CogvideoX-5B as the cloud model and the CogvideoX-2B \cite{yang2024cogvideox} as the edge model. Then we compare EC-Diff against HybridSD \cite{yan2024hybrid} in both tasks, where the edge model inference steps $k$ of HybridSD is set to 10 and 5 based on the configurations from its original experiments.

\noindent{\textbf{Benchmark Datasets.}} Following \cite{yan2024hybrid}, we use 30K prompts from the zero-shot MS-COCO 2014 \cite{lin2014microsoft} validation split to evaluate the results on T2I task. For the T2V task, we follow \cite{zhao2024real} to evaluate the results using all the prompts in VBench \cite{huang2024vbench}.

\noindent{\textbf{Evaluation Metrics.}} For all tasks, we evaluate our proposed method in both quality and efficiency. We report latency and speedup ration to verify the efficiency. For the image generation task, we follow previous works \cite{ye2024training,ma2024deepcache} and evaluate image quality using the following metrics: Peak Signal-to-Noise Ration (PSNR), Learned Perceptual Image Patch Similarity (LPIPS), and Structural Similarity Index Measure (SSIM). For the video generation task, in addition to using the above metrics, we follow previous works \cite{zhao2024real,lv2024fastercache} to compute the VBench metrics to evaluate the performance of the video in multiple dimensions. The details of evaluation metrics are presented in Appendix \ref{sec:metrics} .

\noindent{\textbf{Implementation Details.}} We conduct all experiments on single NVIDIA A800 80G GPU. The sampling timesteps T in the image and video generation task are set to 50 and we use DDIM \cite{song2020denoising} as the sampler. The details and results for the two-stage greedy search under different combinations of edge-cloud models are shown in Appendix \ref{sec:search_result} .

\begin{figure*}[ht]
	\centering
	\includegraphics[width=1\textwidth]{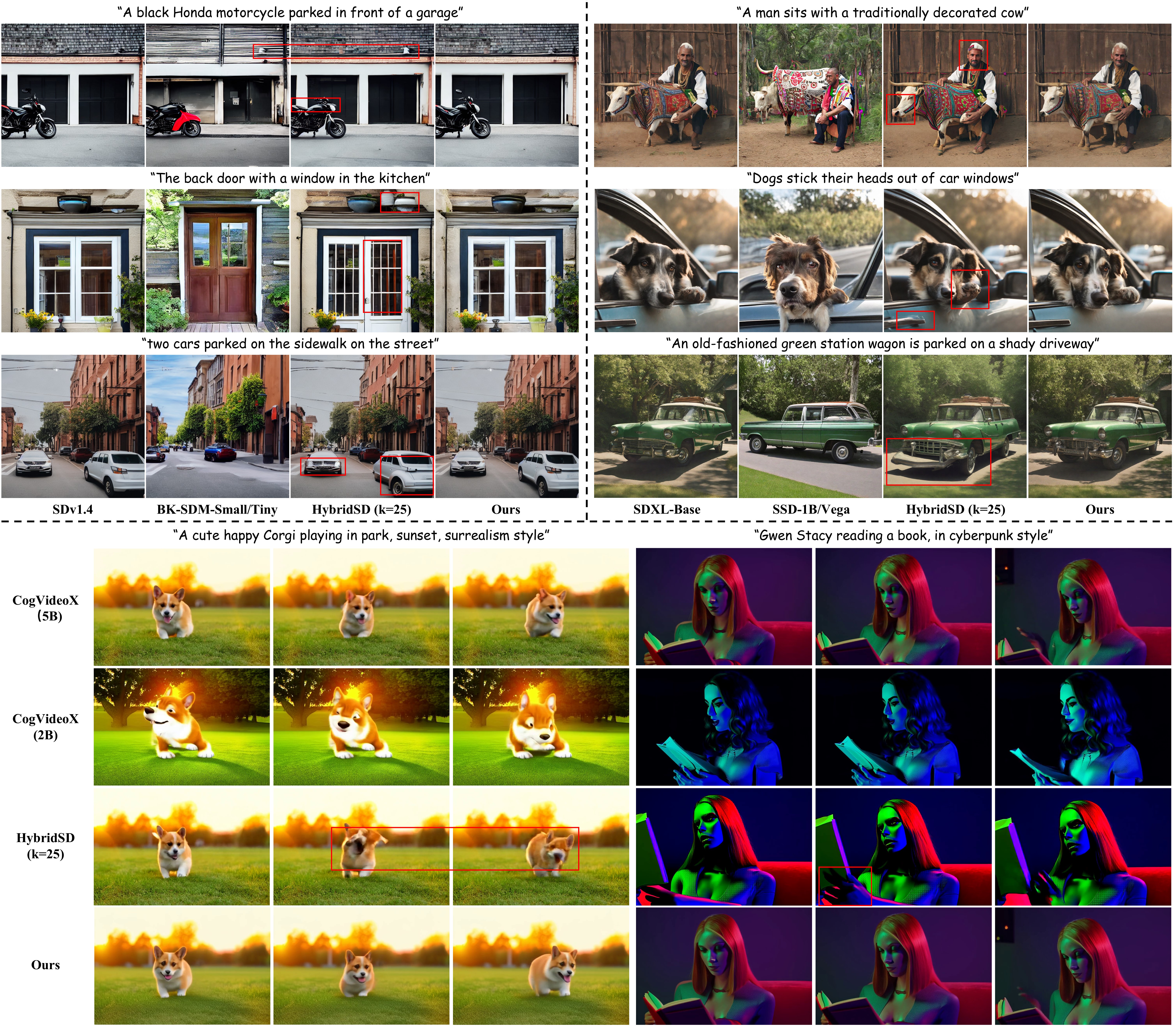}
	\vspace{-0.5cm}
	\caption{Qualitative results of T2I and T2V generation task using different models. To facilitate a better comparison of the improvements in generation quality, we present the generation results from both the cloud and edge models.}
	\label{fig:quality}
\end{figure*}

\subsection{Quantitative Results}
The results are presented in \autoref{tab:quantitive}. Our results indicate that the generation quality of EC-Diff is nearly identical to HybridSD (k=10), which ensures high fidelity to cloud-only inference by executing a higher number of cloud inference steps. However, this approach results in minimal acceleration, with a modest 1.1$\times$ speedup. In contrast, our method achieves an average of 2$\times$ acceleration without compromising generation quality.

Furthermore, our model significantly outperforms HybridSD (k=5) in terms of both quality and speed. Specifically, the PSNR for the T2I task improves by 20\% while for the T2V task increases by 30\%. Notably, the speedup of our method also surpasses that of HybridSD (k=5), which achieves a maximum acceleration of 1.5$\times$, whereas our approach demonstrates a substantial 2.4$\times$ acceleration. These results highlight the superior efficiency of our model, achieving both higher quality and faster generation compared to existing methods. More detailed results are shown in Appendix \ref{sec:more_result}.

\subsection{Qualitative Results}
\subsubsection{\textbf{Generation Comparisons.}} 

\begin{figure*}[ht]
	\centering
	\includegraphics[width=0.95\textwidth]{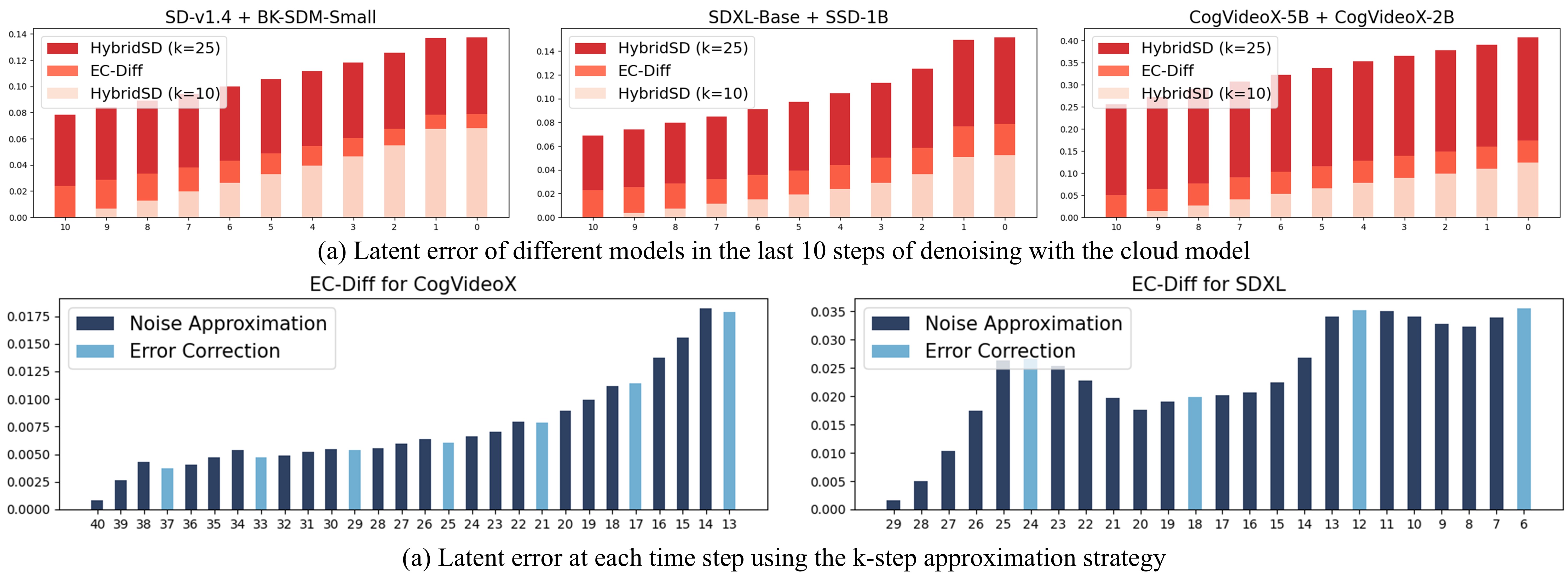}
	\caption{Visualization of latent error under different models}
	\vspace{-0.2cm}
	\label{fig:error}
\end{figure*}

As shown in \autoref{fig:quality}, our model demonstrates superior generation quality and stability. For the image generation task,  Hybrid (k=25) exhibits notable semantic distortions due to reduced cloud inference steps, manifesting as structural anomalies in vehicles and degradation of facial features. In contrast, our model produces results visually consistent with cloud model outputs. For the video generation task, as video content maintain both spatial and temporal consistency, HybridSD (k=25) demonstrates notable instability between frames. For instance, when generating video sequences depicting a puppy running from left to right, Hybrid produces frames with sudden facial distortions, disrupting the visual continuity. In contrast, our method preserves content consistency across sequential frames, ensuring smooth temporal transitions while maintaining semantic integrity throughout the generated video sequence. We recommend watching more qualitative results\footnote{\url{https://ec-diff.github.io/}}.

\subsubsection{\textbf{Denoising error Comparisons.}} As shown in \autoref{fig:error} (a), during the last ten steps of denoising, both our model and HybridSD with more cloud inference (k = 10) maintain tiny latent error, indicating that the cumulative error impact from our k-step noise approximation strategy is small and does not bring visual loss to the generation quality. \autoref{fig:error} (b) shows the two distinct error correction modes of our method. The first mode on the left side demonstrates a clear decrease in latent error immediately following the error correction step. The second mode on the right side shows that the updated noise gradient subsequently facilitates error reduction during the following approximation steps. Both phenomena provide evidence that our method effectively leverages the model's inherent error correction capabilities to maintain high generation quality throughout the denoising process.

\begin{figure}[t!]
\centering
\begin{minipage}[t]{0.48\textwidth}
  \vspace{0pt}
  \centering
  \small
  \setlength{\tabcolsep}{4pt}
    \begin{tabular}{m{3.28cm}ccc}
    \toprule
    Ablation & {PSNR$\uparrow$} & {LPIPS$\downarrow$} & {SSIM$\uparrow$} \\
    \midrule
    Ours          & 29.279 & 0.057 & 0.901 \\
    \midrule
    correct errors twice & 29.278 & 0.055 & 0.903 \\
    w/o $\alpha$  & 8.328  & 0.783 & 0.203 \\
    w/ Euler Scheduler & 29.387 & 0.051 & 0.903 \\
    w/ DPM++ Scheduler~\cite{lu2022dpm} & 28.814 & 0.064 & 0.897 \\
    \bottomrule
    \end{tabular}
  \captionof{table}{Ablation study for our components.}
  \label{tab:ablation}
\end{minipage}
\hfill
\begin{minipage}[t]{0.48\textwidth}
  \vspace{0pt}
  \centering
  \includegraphics[width=0.8\linewidth]{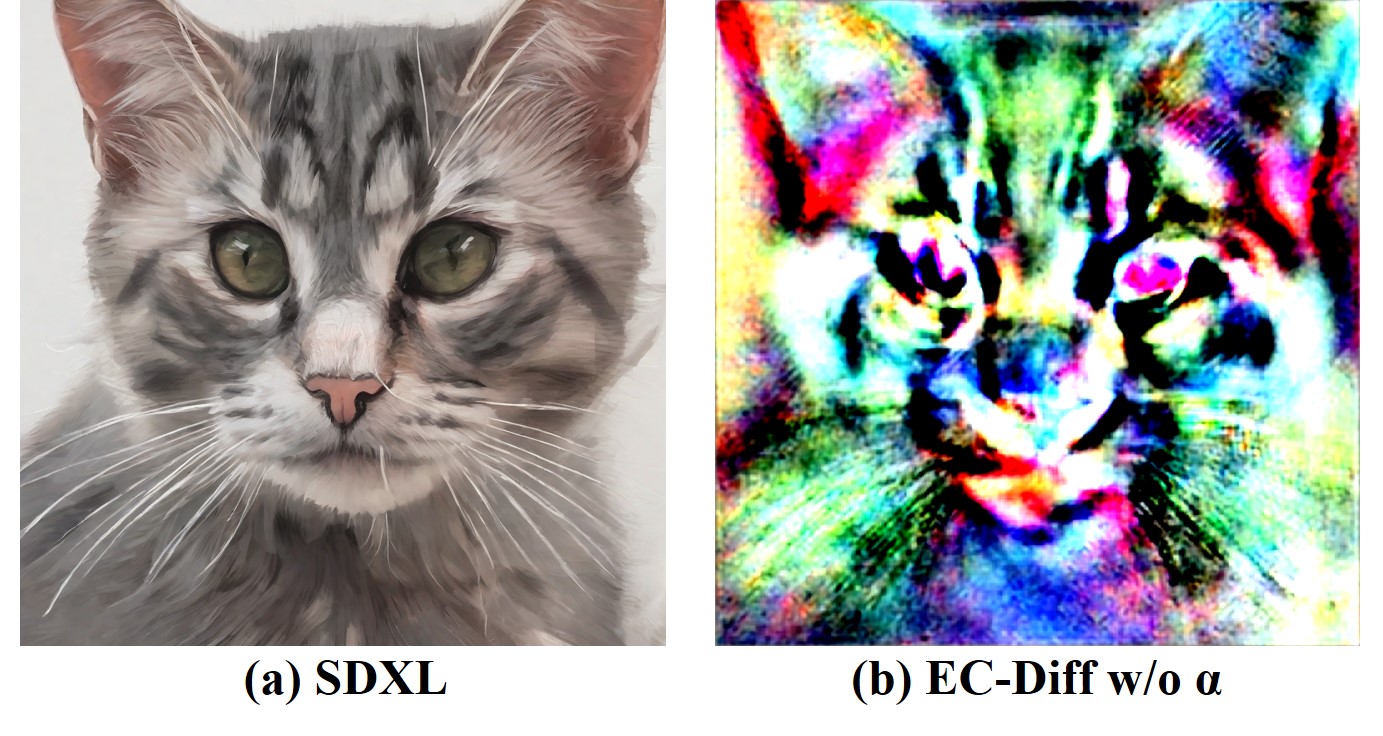}
  \vspace{-0.15cm}
  \captionof{figure}{Generation result w/o $\alpha$}
  \label{fig:ablation}
\end{minipage}%
\end{figure}

\subsection{Ablation Study}

As shown in \autoref{tab:ablation}, the qualitative metrics do not change significantly when correcting errors twice, demonstrating that a single error correction is sufficient for our approach. Noteably, we observe that without smoothing noise gradient leads to a substantial decrease in quality metrics, with anomalies in the generated results as shown in \autoref{fig:ablation}. This degradation occurs because error correction introduces considerable changes in the noise gradient, and without smoothing, these abrupt changes propagate significant errors throughout the denoising process. Additionally, our experiments across different schedulers reveal minimal variation in metrics, indicating that our method maintains consistent performance regardless of the sampling algorithm employed.

\vspace{-0.15cm}
\section{Conclusion}
\vspace{-0.25cm}

In this work, we propose EC-Diff, a novel framework for fast and high-quality diffusion inference via edge-cloud collaboration. The k-step noise approximation strategy enables lossless acceleration of cloud denoising by exploiting the error correction capability of the model, while the two-stage greedy algorithm identifies the optimal cloud-edge switching point to ensure high-quality generation with minimal resource overhead. Experiments are conducted on both image and video diffusion models, showing a good trade-off between high generation quality and low inference cost. We believe this novel approach can inspire future research toward faster and more device-adaptive methods. By leveraging advanced distillation techniques, lightweight models tailored for denoising quality enhancement tasks can be deployed on resource-constrained devices such as mobile phones, enabling broader user to experience the impressive generation result with low latency.

{
    \small
    \bibliographystyle{plain}
    \bibliography{main}
}

\clearpage
\appendix
\section{Appendix}

\subsection{Algorithm of k-step noise approximation strategy}

\begin{algorithm}[H]
	\caption{K-Step Noise Approximation Strategy.}
	\label{alg:k-step}
	\renewcommand{\algorithmicensure}{\textbf{Input:}}
	\begin{algorithmic}[1]
		\Ensure{Cloud Noise Prediction Model $\epsilon_\theta$, Sampling Scheduler $\phi$, Sample Step $T$, Conditional embedding $c$, Pre-Inference step $i$, Approximation Step $k$, Smoothing Factor $\alpha$, Switching Point $s$;}
		\State Initialize Random Noise $x$, Noise Gradient $\delta \epsilon = $ None, Previous Noise List $P_{\text{noise}}=$[].
		\For{$t=T$ to $T - i$}
		\State $P_{\text{noise}}$.append($\epsilon_\theta(x,t,c)$);
		\State Compute $\phi(x,\epsilon_\theta(x,t,c))$ by Eq. (\ref{eq:eq_1});
		\EndFor
		\State $\delta \epsilon \leftarrow P_{\text{noise}}[-1] - P_{\text{noise}}[-2]$ ;
		\While{True}
		\For{$j = 1$ to $k$}
		\State $\epsilon_\theta^{'}(x, t-j,c) \leftarrow P_{\text{noise}}[-1]+ \delta \epsilon$;
		\State $P_{\text{noise}}$.append($\epsilon_\theta(x, t-j,c)^{'}$);
		\State Compute $\phi(x,\epsilon_\theta(x, t-j,c)^{'})$ by Eq. (\ref{eq:eq_1});
		\EndFor
		\State $t \leftarrow t-k-1$;
		\State $P_{\text{noise}}$.append($\epsilon_\theta(x, t, c)$);
		\State Compute $\phi(x,\epsilon_\theta(x, t,c))$ by Eq. (\ref{eq:eq_1});
		\State $\delta \epsilon = \alpha \times (P_{\text{noise}}[-1] - P_{\text{noise}[-2]});$
		\If{$t\le s$}
		\State break;
		\EndIf
		\EndWhile\\
		\Return{$x$.}
	\end{algorithmic}
\end{algorithm}

\subsection{Error Estimation Induced by K-Step Noise Approximation Strategy}

\subsubsection{\textbf{A generalized formula for model inference k steps.}} Suppose we perform one inference from the latent $x_{i-1}$ obtained from the noise $\epsilon_\theta(x_i, i, c)$, the formula is as follows:
\begin{equation}\label{eq:first_inference}
	\small
	x_{i-2}=f(i-2) \cdot x_{i-1}-g(i-2) \cdot \epsilon_\theta(x_{i-1}, i-1, c).
\end{equation}

Similarly, we can obtain the following formula after the second sampling and substituting \autoref{eq:first_inference}:
\begin{equation}\label{eq:second_use_predict}
	\small
	\begin{aligned}
		x_{i-3} &= f(i-3) \cdot x_{i-2}-g(i-3) \cdot \epsilon_\theta(x_{i-2}, i-2, c)
		\\
		&= f(i-3) \cdot f(i-2) \cdot x_{i-1} - f(i-3) \cdot g(i-2) \cdot \epsilon_\theta(x_{i-1}, i-1, c)
		\\
		&\,\,\,\,\, -g(i-3) \cdot \epsilon_\theta(x_{i-2}, i-2, c).
	\end{aligned}
\end{equation}

By analogy, we can obtain the following results after sampling k times using the predicted noise:
\begin{equation}\label{eq:k_inference}
	\small
	\begin{aligned}
		x_{i-1-k} &= \prod\limits_{{\rm{j}} = i - k - 1}^{i - 2} {f(j) \cdot {x_{i - 1}}} - g(i-k-1) \cdot \epsilon_\theta(x_{i-k}, i-k, c)
		\\
		&\,\,\,\,\, - \sum\limits_{m = 0}^{k - 2}{\left(\left(\prod\limits_{j = i-1-k}^{i-3-m}{f(j)}\right)\cdot g(i - 2 - m) \cdot \epsilon_\theta(x_{i-1-m}, i-1-m, c)\right)}.
	\end{aligned}
\end{equation}

\subsubsection{\textbf{Error accumulation during the first application of the k-step noise approximation strategy.}}\label{sec:first_apply_kstep} After initializing the noise gradient, the model performs k sampling steps using approximated noise, followed by one sampling step based on the model-predicted noise. At this point, the accumulated error depends on the model’s ability to correct errors. Here, we provide the detailed proof of this statement.
\begin{figure}[t]
	\centering
	\includegraphics[width=0.8\textwidth]{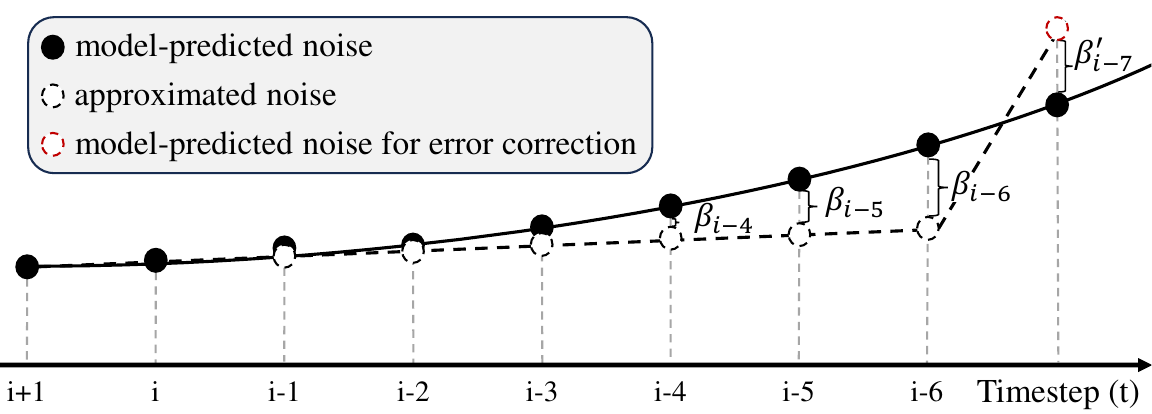}
	\caption{Error at each timestep during the first application of the k-step noise approximation strategy.} 
	\label{fig:proof_1}
	\vspace{-0.5cm}
\end{figure}
\\
\textbf{Assumptions.} As shown in \autoref{fig:proof_1}, we use $\beta$ to represent the error accumulation, and make the assumptions that:\\
(1) $\beta_i$ denotes the error between the approximated noise and the model-predicted noise obtained at step $i$;\\
(2) $\beta_i^{'}$ denotes the amount of noise that the model over-predicts at step $i$ compared to the normal case due to error-correction ability.\\ 
\textbf{Proof.} Taking the example that the model pre-inferences $i$ steps and get $x_{i-1}$, the first time the approximated noise is $\epsilon_\theta(x_{i-1}, i-1)-\beta_{i-1}$, then we can get the following formulation:
\begin{equation}\label{eq:first_use_predict}
	\small
	x_{i-2}^{'}=f(i-2) \cdot x_{i-1}-g(i-2) \cdot (\epsilon_\theta(x_{i-1}, i-1, c)-\beta_{i-1}).
\end{equation}

Similarly, we can obtain the following formula after the second sampling using the approximated noise and substituting \autoref{eq:first_use_predict}:
\begin{equation}\label{eq:second_use_predict}
	\small
	\begin{aligned}
		x_{i-3}^{'} &= f(i-3) \cdot x_{i-2}^{'}-g(i-3) \cdot (\epsilon_\theta(x_{i-2}, i-2, c)-\beta_{i-2})
		\\
		&= f(i-3) \cdot f(i-2) \cdot x_{i-1} - f(i-3) \cdot g(i-2) \cdot \epsilon_\theta(x_{i-1}, i-1,c)
		\\
		&\,\,\,\,\, +(f(i-3) \cdot g(i-2) \cdot \beta_{i-1}+g(i-3) \cdot \beta_{i-2})
		\\
		&\,\,\,\,\, -g(i-3) \cdot \epsilon_\theta(x_{i-2}, i-2, c).
	\end{aligned}
\end{equation}

By analogy, we can obtain the following results after sampling k times using the prediction noise:
\begin{equation}\label{eq:k_use_predict}
	\small
	\begin{aligned}
		x_{i-1-k}^{'} &= \prod\limits_{{\rm{j}} = i - k - 1}^{i - 2} {f(j) \cdot {x_{i - 1}}} - g(i-k-1) \cdot \epsilon_\theta(x_{i-k}, i-k,c)
		\\
		&\,\,\,\,\, - \sum\limits_{m = 0}^{k - 2}{\left(\left(\prod\limits_{j = i-1-k}^{i-3-m}{f(j)}\right)\cdot g(i - 2 - m) \cdot \epsilon_\theta(x_{i-1-m}, i-1-m, c)\right)}
		\\
		&\,\,\,\,\, + \underbrace{\sum\limits_{m = 0}^{k - 2}{\left(\left(\prod\limits_{j = i-1-k}^{i-3-m}{f(j)}\right)\cdot g(i - 2 - m)\cdot \beta_{i-1-m}\right)} + g(i-1-k)\cdot \beta _{i-k}}_{cumulative\,\,\,error}.
	\end{aligned}
\end{equation}

It can be seen that there is an additional error accumulation term on top of \autoref{eq:k_inference} obtained by normal denoising. This term is due to the error between the approximated noise and the model-predicted noise, which accumulates as the number of sample steps increases. 

Next, we calculate the accumulation of errors after denoising with the model-predicted noise for one step and obtain the following formula:

\begin{equation}\label{eq:first_model_inference}
	\small
	\begin{aligned}
		x_{i-k-2}^{'} &= f(i-k-2) \cdot x_{i-k-1}^{'} 
		\\
		&\,\,\,\,\, - g(i-k-2) \cdot (\epsilon_\theta(x_{i-k-1}, i-k-1, c)+\beta_{i-k-1}^{'})
		\\
		&= \prod\limits_{{\rm{j}} = i - k - 2}^{i - 2} {f(j) \cdot {x_{i - 1}}} - g(i-k-2) \cdot \epsilon_\theta(x_{i-k-1}, i-k-1,c)
		\\
		&\,\,\,\,\, - \sum\limits_{m = 0}^{k - 1}{\left(\left(\prod\limits_{j = i-k-2}^{i-3-m}{f(j)}\right)\cdot g(i - 2 - m) \cdot \epsilon_\theta(x_{i-1-m}, i-1-m,c)\right)}
		\\
		&\,\,\,\,\, + \sum\limits_{m = 0}^{k - 1}{\left(\left(\prod\limits_{j = i-k-2}^{i-3-m}{f(j)}\right)\cdot g(i - 2 - m)\cdot \beta_{i-1-m}\right)} 
		\\
		&\,\,\,\,\, - g(i-k-2) \cdot \beta_{i-k-1}^{'}.
	\end{aligned}
\end{equation}

Then we can get the cumulative error equation as follows:
\begin{equation}\label{eq:first_model_inference_error}
	\small
	\begin{aligned}
		\Delta x_{i-k-2} &= x_{i-k-2}^{'} - x_{i-k-2}
		\\
		&= \sum\limits_{m = 0}^{k - 1}{\left(\left(\prod\limits_{j = i-k-2}^{i-3-m}{f(j)}\right)\cdot g(i - 2 - m)\cdot \beta_{i-1-m}\right)}
		\\
		&\,\,\,\,\, - g(i-k-2) \cdot \beta_{i-k-1}^{'}
		\\
		&< \sum\limits_{m = 0}^{{\rm{k - }}1} {{\beta _{i - 1 - m}}}  - \beta _{i - k - 1}^{'} = O(\sum\limits_{m = 0}^{{\rm{k - }}1} {{\beta _{i - 1 - m}}}  - \beta _{i - k - 1}^{'}).
	\end{aligned}
\end{equation}

Since the coefficients $f$ and $g$ are less than or equal to 1, the upper bound of error accumulation of the first application of the $k$-step noise approximation strategy is determined by the error accumulation introduced by the approximated noise and the steps of approximations, and the amount of noise correction applied by the model. When the model's error correction capability is sufficient, it will be able to identify the previously accumulated error noise, making the denoised latent return to the normal condition.

\subsubsection{\textbf{Error accumulation in subsequent strategy cycles.}}\label{sec:second_apply_kstep}
After the first application of the strategy and continues to denoise $k$ steps using the approximated noise, the accumulation of error depends on the noise gradient smoothing coefficient, the steps of approximations and the model's ability to correct errors.  Here, we provide the detailed proof of this statement.
\begin{figure}[ht]
	\centering
	\includegraphics[width=0.7\textwidth]{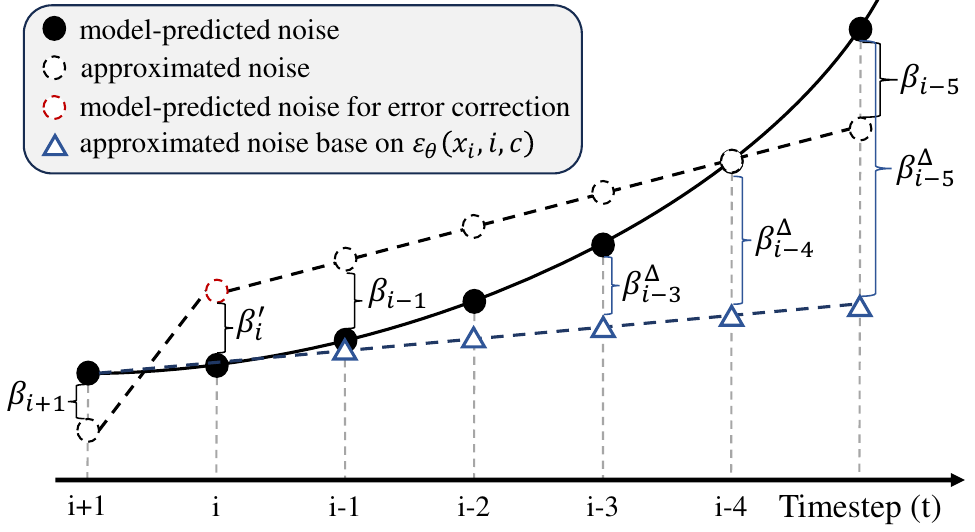}
	\caption{Error at each timestep in subsequent strategy cycles.} 
	\label{fig:proof_2}
\end{figure}
\\
\textbf{Assumptions.} We assume that the latent obtained after the first application of the strategy is $x_{i-1}$.\\
\textbf{Proof.} According to the noise gradient update formula, we use the new noise gradient formula as follows:
\begin{equation}\label{eq:new_noise_direct}
	\small
	\begin{aligned}
		\delta \epsilon &= \alpha \cdot \left((\epsilon_\theta(x_{i}, i,c)+\beta_{i}^{'}) - (\epsilon_\theta(x_{i+1}, i+1,c)- \beta_{i+1})\right)
		\\
		&=\alpha \cdot \left(\epsilon_\theta(x_{i}, i,c)-\epsilon_\theta(x_{i+1}, i+1,c)+\beta_{i}^{'}+\beta_{i+1}\right),
	\end{aligned}
\end{equation}

Next, as shown by the triangular path in \autoref{fig:proof_2}, we use the noise gradient computed based on $x_i$ and $x_{i+1}$ to represent the model-predicted noise, which can be formulated as:
\begin{equation}\label{eq:infer_direct}
	\small
	\begin{aligned}
		\epsilon_\theta(x_{i-k}, i-k,c) &= \epsilon_\theta(x_{i}, i,c)+ \beta^{\triangle}_{i-k}
		\\
		&\,\,\,\,\, +k\cdot (\epsilon_\theta(x_{i}, i,c) - \epsilon_\theta(x_{i+1}, i+1,c)).
	\end{aligned}
\end{equation}

The latent obtained once using model-predicted noise is as follows:
\begin{equation}\label{eq:infer_direct_one}
	\small
	\begin{aligned}
		x_{i-2}^{True} &= f(i-2) \cdot x_{i-1} - g(i-2) \cdot \epsilon_\theta(x_{i-1}, i-1,c)
		\\
		&= f(i-2) \cdot x_{i-1} - g(i-2) \cdot ( 2\cdot \epsilon_\theta(x_{i}, i,c)-\epsilon_\theta(x_{i+1}, i+1,c)
		\\
		&\,\,\,\,\, +\beta^{\triangle}_{i-1}).
	\end{aligned}
\end{equation}

The latent sampled by using approximated noise is as follows:
\begin{equation}\label{eq:infer_predict_one}
	\small
	\begin{aligned}
		x_{i-2} &= f(i-2) \cdot x_{i-1} - g(i-2) \cdot (\epsilon_\theta(x_{i}, i,c)+\beta_{i}^{'}
		\\
		&\,\,\,\,\, +\alpha\cdot \left(\epsilon_\theta(x_{i}, i,c)-\epsilon_\theta(x_{i+1}, i+1,c)+\beta_{i}^{'}+\beta_{i+1}\right)).
	\end{aligned}
\end{equation}

We can obtain the error accumulated at step i-2 as follows:
\begin{equation}\label{eq:error_one}
	\small
	\begin{aligned}
		\Delta x_{i-2} &= x_{i-2} - x_{i-2}^{True}
		\\
		&= -g(i-2) \cdot (\epsilon_\theta(x_{i}, i,c) + \beta_{i}^{'}+ \alpha \cdot (\epsilon_\theta(x_{i}, i,c) - \epsilon_\theta(x_{i+1}, i+1,c) 
		\\
		&\,\,\,\,\, + \beta_{i}^{'}+\beta_{i+1}) - (2 \cdot \epsilon_\theta(x_{i}, i,c) - \epsilon_\theta(x_{i+1}, i+1,c) + \beta^{\triangle}_{i-1}))
		\\
		&= -g(i-2)\cdot((1+\alpha) \cdot \beta_{i}^{'} + \alpha \cdot \beta_{i+1} - \beta^{\triangle}_{i-1} 
		\\
		&\,\,\,\,\, + (\alpha - 1) \cdot (\epsilon_\theta(x_{i}, i,c) - \epsilon_\theta(x_{i+1}, i+1,c))).
	\end{aligned}
\end{equation}

Similarly, we can get the results of the two steps of inference by using model-predicted noise as follows:
\begin{equation}\label{eq:eq:infer_direct_two}
	\small
	\begin{aligned}
		x_{i-3}^{True} &= f(i-3) \cdot x_{i-2}^{True} - g(i-3) \cdot \epsilon_\theta(x_{i-2}, i-2,c)
		\\
		&= f(i-3) \cdot  x_{i-2}^{True} - g(i-3) \cdot ( 3\cdot \epsilon_\theta(x_{i}, i,c)
		\\
		&\,\,\,\,\, -2\cdot \epsilon_\theta(x_{i+1}, i+1,c)+\beta^{\triangle}_{i-2}).
	\end{aligned}
\end{equation}

The result of using the approximated noise twice is as follows:
\begin{equation}\label{eq:eq:infer_predict_one}
	\small
	\begin{aligned}
		x_{i-3} &= f(i-3) \cdot x_{i-2} - g(i-3) \cdot (\epsilon_\theta(x_{i}, i,c)+\beta_{i}^{'}
		\\
		&\,\,\,\,\, +2\alpha\cdot \left(\epsilon_\theta(x_{i}, i,c)-\epsilon_\theta(x_{i+1}, i+1,c)+\beta_{i}^{'}+\beta_{i+1}\right)).
	\end{aligned}
\end{equation}

We can obtain the error accumulated at step $i-3$ as follows:
\begin{equation}\label{eq:error_two}
	\small
	\begin{aligned}
		\Delta x_{i-3} &= x_{i-3} - x_{i-3}^{True}
		\\
		&= f(i-3) \cdot \delta x_{i-2} - g(i-3) \cdot (\epsilon_\theta(x_{i}, i,c) + \beta_{i}^{'} 
		\\
		&\,\,\,\,\, +2\alpha \cdot (\epsilon_\theta(x_{i}, i,c) - \epsilon_\theta(x_{i+1}, i+1,c) + \beta_{i}^{'}+\beta_{i+1})
		\\
		&\,\,\,\,\,  - (3\cdot \epsilon_\theta(x_{i}, i,c) - 2\cdot \epsilon_\theta(x_{i+1}, i+1,c) + \beta^{\triangle}_{i-2}))
		\\
		&=f(i-3) \cdot \delta x_{i-2} -g(i-3)\cdot((1+2\alpha) \cdot \beta_{i}^{'} + 2\alpha \cdot \beta_{i+1} - \beta^{\triangle}_{i-2} 
		\\
		&\,\,\,\,\, + (2\alpha - 2) \cdot (\epsilon_\theta(x_{i}, i,c) - \epsilon_\theta(x_{i+1}, i+1,c)))
		\\
		&= -((f(i-3) \cdot g(i-2) \cdot (1+\alpha) + g(i-3) \cdot (1+2\alpha))\cdot \beta_{i}^{'}
		\\
		&\,\,\,\,\, + (f(i-3) \cdot g(i-2) \cdot \alpha + g(i-3) \cdot 2\alpha)\cdot \beta_{i+1}
		\\
		&\,\,\,\,\, - (f(i-3) \cdot g(i-2) \cdot \beta^{\triangle}_{i-1} + g(i-3) \cdot \beta^{\triangle}_{i-2})
		\\
		&\,\,\,\,\, +(f(i-3) \cdot g(i-2) \cdot (\alpha -1) + g(i-3) \cdot (2\alpha -2))
		\\
		&\,\,\,\,\, \cdot (\epsilon_\theta(x_{i}, i,c) - \epsilon_\theta(x_{i+1}, i+1,c))).
	\end{aligned}
\end{equation}

By analogy, we can obtain the following results after sampling k times using the prediction noise:
\begin{equation}\label{eq:error_k}
	\small
	\begin{aligned}
		\Delta x_{i-1-k} &= (\sum\limits_{m = 1}^{k - 1} {((\prod\limits_{j = m + 1}^k {f(i - 1 - j)} )\cdot g(i - 1 - m)\cdot \beta^{\triangle}_{i-m}} ) 
		\\
		&\,\,\,\,\,\,\,\,\,\,\,\,\, + g(i - 1 - k)\cdot \beta^{\triangle}_{i-k})
		\\
		&\,\,\,\,\, -(\sum\limits_{m = 1}^{k - 1} {((\prod\limits_{j = m + 1}^k {f(i - 1 - j)} )\cdot g(i - 1 - m)\cdot (1 + m\cdot \alpha )} ) 
		\\
		&\,\,\,\,\,\,\,\,\,\,\,\,\, + g(i - 1 - k)\cdot (1 + k\cdot \alpha ))\cdot\beta _i^{'}
		\\
		&\,\,\,\,\, -(\sum\limits_{m = 1}^{k - 1} {((\prod\limits_{j = m + 1}^k {f(i - 1 - j)} )\cdot g(i - 1 - m)\cdot m \cdot \alpha} ) 
		\\
		&\,\,\,\,\,\,\,\,\,\,\,\,\, + g(i - 1 - k)\cdot k \cdot \alpha )\cdot\beta _{i+1}
		\\
		&\,\,\,\,\, -(\sum\limits_{m = 1}^{k - 1} {((\prod\limits_{j = m + 1}^k {f(i - 1 - j)} )\cdot g(i - 1 - m)\cdot (m \cdot \alpha -m )} ) 
		\\
		&\,\,\,\,\,\,+ g(i - 1 - k)\cdot (k \cdot \alpha-k) )\cdot(\epsilon_\theta(x_{i}, i,c) - \epsilon_\theta(x_{i+1}, i+1,c))
		\\
		&< \sum\limits_{m = 1}^k {{\beta^{\triangle}_{i - m}}}  - \sum\limits_{m = 1}^k {(1 + m\alpha )\beta _i^{'}}  - \sum\limits_{m = 1}^k {m\alpha {\beta _{i + 1}}} 
		\\
		&= \sum\limits_{m = 1}^k {({\beta^{\triangle}_{i - m}} - (1 + m\alpha )\beta _i^{'} - m\alpha {\beta _{i + 1}})} 
		\\
		&=O(\sum\limits_{m = 1}^k {({\beta^{\triangle}_{i - m}} - (1 + m\alpha )\beta _i^{'} - m\alpha {\beta_{i + 1}})});
	\end{aligned}
\end{equation}

After a single model inference step to correct the error, the remaining accumulated error depends on whether the cumulative error in \autoref{eq:error_k} is within the model's correction capacity. If it is, the remaining error is zero; otherwise, the error will accumulate into the next strategy cycle. 
This process is repeated, with the error accumulation in \autoref{eq:error_k} alternating with model correction, resulting in a cyclical process of error accumulation and mitigation.

\subsection{Two-Stage Greedy Search Algorithm}

The parameters $k$ and $\alpha$ in the $k$-step noise approximation strategy affect the error accumulation and generation speed in the cloud model's inference phase, while the cloud-edge switching point determines whether the edge model can follow the cloud model's semantic planning. Considering that the error accumulation in the cloud model influences the choice of the cloud-edge switching point, we adopt a two-stage greedy search algorithm. 

The first stage seeks the optimal $(k,\alpha)$ combination for the $k$-step noise approximation strategy used in the cloud model, ensuring the best generation quality and speed with minimal error accumulation, thus minimizing its impact on the cloud-edge switching point. 
The second stage solves for the optimal cloud-edge switching point given the best $(k,\alpha)$ combination. Both stages use a greedy algorithm with an early stopping mechanism. Specifically, we track the impact of parameter changes on the objective function. If the change is beneficial to the objective function, we continue exploring in that direction; otherwise, we explore in the opposite direction. If no improvement is observed after three consecutive parameter adjustments, we consider the optimal solution found and terminate the search. The complete two-stage greedy search algorithm is summarized in Algorithm \autoref{alg:greedy_1} and Algorithm \autoref{alg:greedy_2}.
\begin{algorithm}[H]
	\caption{Greedy Search Algorithm Stage-1.}
	\label{alg:greedy_1}
	\renewcommand{\algorithmicensure}{\textbf{Input:}}
	\begin{algorithmic}[1]
		\Ensure{Cloud Model $M$ With K-Step Approximation Strategy, Dataset $D$, Evaluate Function $E$, Pre-Inference Step $p$.}
		\State Initialize $V_\text{best}=0$, $k_\text{best}=0$, $\alpha_\text{best}=0$, decreasing\_count=0, $a$\_value=[0.1,0.2,0.3,0.4,0.5,0.6,0.7,0.8,0.9];
		\For{$k=1$ to $5$}
		\State Initialize $V^{'}_{best}=0$,$\alpha^{'}_{best}=0$, d=1,$c_{\text{worse}}=0$,vis=[],$\alpha$=0.5;
		\State vis.append($\alpha$);
		\State objective\_value = $E$($M$,$D$,$p$,$k$,$\alpha$);
		\State $V^{'}_{best}=$objective\_value;
		\State $\alpha^{'}_{best}$=$\alpha$;
		\While{$c_{\text{worse}}$<3 and length(vis)<length($\alpha$\_value)}
		\State next\_$\alpha$ = null;
		\If{$d$==1}
		\For{$v$ in $\alpha$\_value}
		\If{$v$ > $\alpha$ and $v$ not in vis}
		\State next\_$\alpha$ = $v$;
		\State break;
		\EndIf
		\EndFor
		\Else
		\For{$v$ in reverse($\alpha$\_value)}
		\If{$v$ < $\alpha$ and $v$ not in vis}
		\State next\_$\alpha$ = $v$;
		\State break;
		\EndIf
		\EndFor
		\EndIf
		\If{next\_$\alpha$ is null}
		\State break;
		\EndIf
		\State $\alpha$ = next\_$\alpha$;
		\State vis.append($\alpha$);
		\State objective\_value=$E$($M$,$D$,$p$,$k$,$\alpha$);
		\If{objective\_value>$V^{'}_{\text{best}}$}
		\State $V^{'}_{\text{best}}$=objective\_value;
		\State $\alpha^{'}_{\text{best}}$ = $\alpha$;
		\State $c_{\text{worse}}$=0;
		\Else
		\State $c_{\text{worse}}$+=1;
		\State $d$ = -$d$;
		\EndIf
		\EndWhile
		\If{$V^{'}_{\text{best}}$>$V_{\text{best}}$}
		\State $V_{\text{best}}$=$V^{'}_{\text{best}}$;
		\State $k_{\text{best}}$=$k$;
		\State $\alpha_{\text{best}}$=$\alpha^{'}_{\text{best}}$;
		\State decreasing\_count=0;
		\Else
		\State decreasing\_count+=1;
		\EndIf
		\If{decreasing\_count$\ge$3}
		\State break;
		\EndIf
		\EndFor
		\\
		\Return{($k_{\text{best}}$,$\alpha_{\text{best}}$).}
	\end{algorithmic}
\end{algorithm}

\begin{algorithm}[H]
	\caption{Greedy Search Algorithm Stage-2.}
	\label{alg:greedy_2}
	\renewcommand{\algorithmicensure}{\textbf{Input:}}
	\begin{algorithmic}[1]
		\Ensure{Edge-Cloud Model $M$ With K-Step Approximation Strategy, Dataset $D$, Evaluate Function $E$, Pre-Inference Step $p$, Noise Approximation Step $k$, Gradient Smooth Factor $\alpha$.}
		\State Initialize $V_\text{best}=0$,  $s_\text{best}=0$, decreasing\_count=0, d=1,$c_{\text{worse}}=0$,vis=[],$s$=35, $s$\_value=[30,31,32,33,34,35,36,37,38,39,40];
		\State vis.append($s$);
		\State objective\_value = $E$($M$,$D$,$p$,$k$,$\alpha$,$s$);
		\State $V_{best}=$objective\_value;
		\State $s_{best}$=$s$;
		\While{$c_{\text{worse}}$<3 and length(vis)<length($s$\_value)}
		\State next\_$s$ = null;
		\If{$d$==1}
		\For{$v$ in $s$\_value}
		\If{$v$ > $s$ and $v$ not in vis}
		\State next\_$s$ = $v$;
		\State break;
		\EndIf
		\EndFor
		\Else
		\For{$v$ in reverse($s$\_value)}
		\If{$v$ < $s$ and $v$ not in vis}
		\State next\_$s$ = $v$;
		\State break;
		\EndIf
		\EndFor
		\EndIf
		\If{next\_$s$ is null}
		\State break;
		\EndIf
		\State $s$ = next\_$s$;
		\State vis.append($s$);
		\State objective\_value=$E$($M$,$D$,$p$,$k$,$\alpha$,$s$);
		\If{objective\_value>$V_{\text{best}}$}
		\State $V_{\text{best}}$=objective\_value;
		\State $s_{\text{best}}$ = $s$;
		\State $c_{\text{worse}}$=0;
		\Else
		\State $c_{\text{worse}}$+=1;
		\State $d$ = -$d$;
		\EndIf
		\EndWhile
		\\
		\Return{$s$.}
	\end{algorithmic}
\end{algorithm}

\subsection{Metrics}
\label{sec:metrics}
In this work, we evaluate our methods using several established metrics to comprehensively assess video quality and image/video similarity. On the one hand, we assess video generation quality by the benchmark VBench, which is well aligned with human perceptions.

\noindent \textbf{VBench.} VBench is a benchmark suite designed for evaluating video generative models, which uses a hierarchical approach to break down 'video generation quality' into various specific, well-defined dimensions. Specifically, VBench comprises 16 dimensions in video generation, including Subject Consistency, Background Consistency, Temporal Flickering, Motion Smoothness, Dynamic Degree, Aesthetic Quality, Imaging Quality, Object Class, Multiple Objects, Human Action, Color, Spatial Relationship, Scene, Appearance Style, Temporal Style, Overall Consistency. In experiments, we adopt the VBench evaluation framework and utilize the official code to apply weighted scores to assess generation quality.

On the other hand, we also evaluate the performance of the edge-cloud collaboration model by the following metrics. We compare the generated images and videos from the cloud model with those from the edge-cloud collaboration model and edge model. For video generation, the metrics are computed on each frame of the video and then averaged over all frames to provide a comprehensive assessment.

\noindent \textbf{Peak Signal-to-Noise Ratio (PSNR)}. PSNR is a widely used metric for measuring the quality of reconstruction in image processing. It is defined as:
\begin{equation}
	\text{PSNR} = 10 \cdot \log_{10} \left(\frac{R^2}{\text{MSE}}\right),
\end{equation}
where $R$ is the maximum possible pixel value of the image and MSE denotes the Mean Squared Error between the reference image and the reconstructed image. Higher PSNR values indicate better quality, as they reflect a lower error between the compared images. For video evaluation, PSNR is computed for each frame and the results are averaged to obtain the overall PSNR for the video. However, PSNR primarily measures pixel-wise fidelity and may not always align with perceived image quality. 

\noindent \textbf{Learned Perceptual Image Patch Similarity (LPIPS).} LPIPS is a metric designed to capture perceptual similarity between images more effectively than pixel-based measures. It is based on deep learning models that learn to predict perceptual similarity by training on large datasets. It measures the distance between features extracted from pre-trained deep networks. The LPIPS score is computed as:
\begin{equation}
	\text{LPIPS} = \sum_{i} \alpha_i \cdot \text{Dist}(F_i(I_1), F_i(I_2)),
\end{equation}
where $F_i$ represents the feature maps from different layers of the network, $ I_1$ and $I_2$ are the images being compared, $\text{Dist}$ is a distance function (often L2 norm), and $\alpha_i$ are weights for each feature layer. Lower LPIPS values indicate higher perceptual similarity between the images, aligning better with human visual perception compared to PSNR. LPIPS is calculated for each frame of the video and averaged across all frames to produce a final score.

\noindent \textbf{Structural Similarity Index Measure (SSIM)}. SSIM evaluate the similarity between two images by considering changes in structural information, luminance, and contrast. SSIM is computed as:
\begin{equation}
	\text{SSIM}(x, y) = \frac{(2 \mu_x \mu_y + C_1)(2 \sigma_{xy} + C_2)}{(\mu_x^2 + \mu_y^2 + C_1)(\sigma_x^2 + \sigma_y^2 + C_2)},
\end{equation}
where $\mu_x$ and $\mu_y$ are the mean values of image patches, $\sigma_x^2$ and $\sigma_y^2$ are the variances, $\sigma_{xy}$ is the covariance, and $C_1$ and $ C_2$ are constants to stabilize the division with weak denominators. SSIM values range from -1 to 1, with 1 indicating perfect structural similarity. It provides a measure of image quality that reflects structural and perceptual differences. For video evaluation, SSIM is calculated for each frame and then averaged over all frames to provide an overall similarity measure.

\begin{table*}[t]
    \centering
    \begin{minipage}[t]{\textwidth}
        \centering
        \vspace{0.2cm}
        \caption{VBench results using different methods under CogvideoX-5B and CogvideoX-2B edge-cloud combination, Part 1.}  
        \label{tab:vbench1}
        \resizebox{1\linewidth}{!}{
            \begin{tabularx}{1.35\textwidth}{>{\centering\arraybackslash}c >{\centering\arraybackslash}p{1.8cm} >{\centering\arraybackslash}p{1.8cm} >{\centering\arraybackslash}p{1.8cm} >{\centering\arraybackslash}p{1.8cm} >{\centering\arraybackslash}p{1.8cm} >{\centering\arraybackslash}p{1.8cm} >{\centering\arraybackslash}p{1.8cm}>{\centering\arraybackslash}p{1.8cm}}
                \toprule
                \textbf{Model} & \textbf{Subject Consistency} & \textbf{Background Consistency} & \textbf{Temporal Flickering} & \textbf{Motion Smoothness} & \textbf{Dynamic Degree} & \textbf{Aesthetic Quality} & \textbf{Imaging Quality} & \textbf{Object Class}\\
                \midrule
                5B & 94.22 & 96.74 & 98.78 & 97.85 & 73.18 & 62.63 & 67.97 & 83.76 \\
                2B & 92.18 & 95.48 & 98.68 & 96.93 & 71.86 & 60.83 & 67.97 & 82.77 \\
                Hb-10 & 94.05 & 96.38 & 98.78 & 97.55 & 73.18 & 62.39 & 67.97 & 83.70 \\
                Hb-25 & 92.89 & 96.01 & 98.73 & 97.01 & 71.98 & 61.13 & 67.97 & 82.85 \\
                Ours & 94.01 & 96.31 & 98.78 & 97.51 & 73.18 & 62.38 & 67.97 & 83.69 \\
                \bottomrule
            \end{tabularx}}
    \end{minipage}
    \begin{minipage}[t]{\textwidth}
        \centering
        \vspace{0.2cm}
        \caption{VBench results using different methods under CogvideoX-5B and CogvideoX-2B edge-cloud combination, Part 2.}  
        \label{tab:vbench2}
        \resizebox{1\linewidth}{!}{
            \begin{tabularx}{1.25\textwidth}{>{\centering\arraybackslash}c >{\centering\arraybackslash}p{1.8cm} >{\centering\arraybackslash}p{1.8cm} >{\centering\arraybackslash}p{1cm} >{\centering\arraybackslash}p{1.8cm} >{\centering\arraybackslash}p{1cm} >{\centering\arraybackslash}p{1.8cm} >{\centering\arraybackslash}p{1.8cm}>{\centering\arraybackslash}p{1.8cm}}
                \toprule
                \textbf{Model} & \textbf{Multiple Objects} & \textbf{Human Action} & \textbf{Color} & \textbf{Spatial Relationship} & \textbf{Scene} & \textbf{Appearance Style} & \textbf{Temporal Style} & \textbf{Overall Consistency}\\
                \midrule
                5B & 65.12 & 87.85 & 85.66 & 70.28 & 35.25 & 23.80 & 24.36 & 26.66 \\
                2B & 62.63 & 85.33 & 80.31 & 67.90 & 35.18 & 22.43 & 24.33 & 25.96 \\
                Hb-10 & 65.12 & 87.08 & 84.59 & 70.23 & 35.18 & 23.80 & 24.36 & 26.59 \\
                Hb-25 & 63.76 & 85.54 & 81.11 & 69.18 & 35.22 & 22.58 & 24.35 & 26.12 \\
                Ours & 65.12 & 87.05 & 84.56 & 70.23 & 35.17 & 23.80 & 24.36 & 26.58 \\
                \bottomrule
            \end{tabularx}}
    \end{minipage}
    \begin{minipage}[t]{\textwidth}
        \centering
        \vspace{0.2cm}
        \caption{Optimal parameter combination results for different edge-cloud model combinations.}
    	\label{tab:greedy_result}
    	\setlength{\tabcolsep}{0.45mm}
    	\begin{tabularx}{\linewidth}{c >{\centering\arraybackslash}X>{\centering\arraybackslash}X>{\centering\arraybackslash}X>{\centering\arraybackslash}X}
    		\toprule
    		\textbf{Model} &  \textbf{$p$} & \textbf{$k$} & \textbf{$\alpha$}  & \textbf{$s$} \\
    		\midrule
    		SD-v1.4 + BK-SDM-Small & 7 & 2 & 0.3 & 34 \\
    		SD-v1.4 + BK-SDM-Tiny & 7 & 2 & 0.3 & 37 \\
    		SDXL-Base + SSD-1B & 6 & 5 & 0.2 & 30 \\
    		SDXL-Base + SSD-Vega & 6 & 5 & 0.2 & 36 \\
    		CogvideoX-5B + CogvideoX-2B & 10 & 3 & 0.2 & 38 \\
    		\bottomrule
    	\end{tabularx}
    \end{minipage}
    \begin{minipage}[t]{\textwidth}
        \centering
        \vspace{0.2cm}
        \caption{Generation time of different models.}
    	\label{tab:speed}
    	\setlength{\tabcolsep}{0.45mm}
    	\begin{tabularx}{\linewidth}{c c>{\centering\arraybackslash}X>{\centering\arraybackslash}X>{\centering\arraybackslash}X}
    		\toprule
    		\textbf{Cloud} &  \textbf{Edge} & \textbf{Hb-10} & \textbf{Hb-25}  & \textbf{Ours} \\
    		\midrule
    		\multirow{2}{*}{SD-v1.4 (1.70)} & BK-SDM-Small (1.18) & 1.61 & 1.39 & 1.16 \\
    		& BK-SDM-Tiny (1.10) & 1.58 & 1.36 & 1.05 \\
    		\multirow{2}{*}{SDXL-Base (6.25)} & SSD-1B (4.29) & 5.87 & 5.28 & 2.77 \\
    		& SSD-Vega (3.27) & 5.64 & 4.76 & 2.61 \\
    		CogvideoX-5B (245) & CogvideoX-2B (91) & 211.17 & 165.95 & 105.61 \\
    		\bottomrule
    	\end{tabularx}
    \end{minipage}
\end{table*}

\subsection{Two-Stage Greedy Search Result}
\label{sec:search_result}

In the objective function, $s^{'}$ is set to 30, $w_1$ is set to 0.3, $w_2$ is set to 0.3, and $w_3$ is set to 0.4. For the image generation model, we randomly selected 100 prompts from the MS-COCO 2014 training set as the dataset for the objective function calculation. For the video generation model, we randomly selected 100 prompts from the WebVid training set as the dataset for the objective function calculation. The optimal parameter combination results for different edge-cloud model combinations are shown in \autoref{tab:greedy_result}.

\subsection{More Quantitative Results}
\label{sec:more_result}
\subsubsection{\textbf{Speedups.}} \autoref{tab:speed} demonstrates the inference time for different models, where Hb stands for HybridSD and the time metric is in seconds. For both Hybrid and our method, a single data upload and download is required. The data upload refers to the process of the user's prompt being transmitted to the cloud. Due to the relatively small data volume involved in this transmission, the associated latency can be considered negligible in this context. The data download refers to the process of transferring the latent and prompt embedding from the cloud server to the edge server. Let $t$ be the transmission time, $D$ denote the data size, and $B$ be the network bandwidth, we have $t = \frac{D}{B}$. We leverage the mean WIFI bandwidth of 18.88Mbps. For SD-v1.4, the latent size is 4$\times$64$\times$64, and prompt embedding size is 2$\times$77$\times$768. The comulative data in FP16 precision is totals 263KB and the cost of transmitting is approximately 0.11s. For SDXL-Base, the latent size is 4$\times$128$\times$128, and prompt embedding size is 2$\times$77$\times$2048. The comulative data in FP16 precision is totals 744KB and the cost of transmitting is approximately 0.32s. For CogVideoX, the latent size is 3$\times$16$\times$60$\times$90, and prompt embedding size is 2$\times$226$\times$4096. The comulative data in FP16 precision is totals 4122KB and the cost of transmitting is approximately 1.75s. The transmission time is only a small part of the overall time. The generation time in \autoref{tab:speed} include the estimated data transfer time.

It can be seen that on the image generation task, the inference time of our model is even less than the inference time of the edge model, which further demonstrates the effectiveness of our model in terms of generation speed and quality.

\subsubsection{\textbf{Detailed results for Vbench.}} We show detailed results for the 16 metrics of Vbench in \autoref{tab:vbench1} and \autoref{tab:vbench2}.


\end{document}